\documentclass[11pt,a4paper]{article}
\usepackage[hyperref]{acl2020}
\usepackage{times}
\usepackage{latexsym}

\usepackage{microtype}

\aclfinalcopy 

\title{Dialog as a Vehicle for Lifelong Learning}

\author{Aishwarya Padmakumar \\
  Department of Computer Science \\
  University of Texas at Austin \\
  \texttt{aish@cs.utexas.edu} \\\And
  Raymond J. Mooney \\
  Department of Computer Science \\
  University of Texas at Austin \\
  \texttt{mooney@cs.utexas.edu} \\}

\date{}

\begin{document}
\maketitle
\begin{abstract}
Dialog systems research has primarily been focused around two main types of applications -- task-oriented dialog systems that learn to use clarification to aid in understanding a goal, and open-ended dialog systems that are expected to carry out unconstrained ``chit chat'' conversations. 
However, dialog interactions can also be used to obtain various types of knowledge that can be used to improve an underlying language understanding system, or other machine learning systems that the dialog acts over. 
In this position paper, we present the problem of designing dialog systems that enable lifelong learning as an important challenge problem, in particular for applications involving physically situated robots.  
We include examples of prior work in this direction, and discuss challenges that remain to be addressed.   
\end{abstract}

\section{Introduction}
Dialog systems research has primarily been focused around two main types of applications -- task-oriented dialog systems that learn to use clarification to aid in understanding a user's goal~\citep{young:procieee13}, and open-ended dialog systems that are expected to carry out unconstrained ``chit chat'' conversations~\citep{serban:aaai2016}. 
Much of this research, assumes access to training dialogs of the type the system is expected to perform, and aims to build a dialog system that can then engage in the same type of interactions. 

This is also the case with most machine learning research, which is focused on a learning problem in the context of a fixed domain and task, that do not change between training and test time.
However, when these systems are used in real-world scenarios, the domain is often wider than that of the original training set, and the requirements of the task may change over time.

Lifelong learning research aims to develop machine learning systems that can be robust to this kind of change, making use of knowledge from previous tasks to improve the performance and sample efficiency of future tasks~\cite{chen:18}.
Lifelong learning can reduce the dependence of learned systems on narrow well-defined tasks and large annotated datasets.
In this position paper, we present the problem of designing dialog systems that enable lifelong learning as an important challenge problem, in particular for applications involving physically situated robots.  

Lifelong learning is particularly relevant for robotics applications that involve an agent physically interacting with its environment because it is difficult and expensive to obtain labelled data during training that adequately covers all scenarios that the agent is likely to encounter during operation. 
Recent work has tried to address this issue using simulation techniques to increase the robustness of differences in the task and domain from training and operation time~\citep{tobin:iros17,tobin:iros18}.
However, a complementary direction would be to leverage interactions with humans that such a system is likely to have during operation to obtain additional labelled data to adapt to changes in domains or tasks that occur during operation. 

Dialog systems by their very nature place the system in a position where it is interacting with a human user. Thus the system is in a position where it can query the user for additional information which may be useful for future interactions.
With dialog systems, as with most machine learning systems, it is reasonable to assume that if the trained dialog system interacts with users in the future, those dialogs can be used to further improve the system.
In particular, since open-ended dialog systems do not in principle make assumptions about the domain of discussion or the types of dialog acts that the user and system engage in, they can be considered to be performing some form of lifelong learning.

However, there are many types of information that dialog systems can explicitly query users for during operation.
Some examples include entries to be added to a knowledge base (for example, a service robot in an office may ask for the name of a new employee), new words that refer to concepts for which the system has a learned model (for example, a dialog agent may clarify the meaning of a word from a different language introduced due to code-switching), or labels that can be used to train supervised models (for example, a home robot may ask whether the new piece of furniture you bought is a type of chair). 

We propose a new focus area for dialog systems research that includes identifying such information-gathering dialog acts that are relevant to different types of dialog systems, learning dialog systems that make use of such dialog acts, and user studies and other supportive research necessary for making such systems more usable in real-world scenarios.
We call this area ``dialog for supporting lifelong learning,'' and present it as an interesting challenge problem for dialog researchers, and review some initial directions on work in this area. 
We believe this is especially relevant for dialog systems on embodied robots, as these systems face more difficulties due to the shortage of available training data, and can hence benefit more by using learning techniques that are better designed to adapt to novel test data.
We believe this area presents a number of interesting challenges regarding dataset and task design, speech processing, sample efficiency, and dialog user analyses. 
The rest of this position paper discusses existing work in the area and challenges for the future.

\section{Prior and Related Work}

Open-domain dialog systems consisting of learnable components that can be improved from dialogs can be considered a form of lifelong learning, since they can in principle learn to adapt to a variety of situations. 
However, these systems are typically designed with the objective of keeping a user engaged in conversation~\citep{cervone:app17}, as opposed to expanding the range of topics the system can converse about.    
There is some work on extracting information such as user attributes from open ended dialogs~\citep{wu:arxiv19} with potential applications in personalized recommendation, but the use of extracted information in such applications is yet to be tested empirically. 
Also related is the generation of curiosity-driven questions -- questions that would enrich the system's knowledge~\citep{scialom:19}. 
However, for practical lifelong learning, it would be additionally desirable to test that such question asking  enables the system to perform better at some downstream task, for example question answering. 

Although task-oriented dialogs are typically more restricted, the information-gathering style questions used in these dialogs can be used for lifelong learning. 
For example, \citet{kollar:icra13} use such queries to explicitly learn a knowledge base of referring expressions for people, locations and actions.
More recently, \citet{she:acl17} combine standard slot-value style clarification queries along with explicit knowledge seeking queries to build a knowledge base of the physical effects of actions on real world objects, while simultaneously using this to plan for and accomplish goals specified by the user.

Information from task-oriented dialogs can also be used to improve natural language understanding. 
\citet{thomason:ijcai15} use the structure of task oriented dialogs, particularly the answers to clarification questions, to obtain weakly supervised training examples to improve a semantic parser. 
This can adapt to some changes in language use over time, as can end-to-end dialog systems~\citep{wen:eacl17}, and those whose language understanding components can be updated over time in other ways from new dialogs~\citep{mesnil:interspeech13}.  
Other work has also shown that the use of clarifications can improve the future performance of an agent at following route instructions in simulated home environments~\citep{chi:20}.

There are also some dialog tasks that are designed solely for teaching specific language understanding capabilities to a system.
For example, \citet{yu:sigdial2016} and \citet{yu:acl2017} focus on learning dialog systems for a task where the system has to learn words for perceptual concepts using dialogs that involve identification and description of objects based on these concepts.

Other dialog frameworks more explicitly try to combine learning of new concepts with using them in an end task, as would be required by an agent performing lifelong learning.
One such framework is Opportunistic Active Learning~\citep{thomason:corl17,padmakumar:emnlp18}, which incorporates active learning queries into an interactive task. 
Instead of having separate a separate phase for model improvement, and a testing phase for using the model in a downstream task, this framework creates dialog tasks in which the system has to learn to trade off model improvement with task completion.  The system looks for opportunities to ask questions about objects in its environment (e.g. ``Would you use the word 'squishy' to describe this object?'') and has been applied to robotics and grounded language domains. These questions may be ``off topic'', that is, not related to the current task, but are expected to be useful for acquiring knowledge that may aid future downstream tasks.
While this work assumes that the underlying models being improved are binary classifiers, other works in learning reinforcement learning policies for selecting active learning queries~\citep{fang:emnlp17,bachman:icml17} indicate how this framework can be generalized beyond the use of binary classifiers. 

Early work in lifelong learning was motivated by control problems in robotics~\citep{thrun:ras95} in order to overcome the difficulties of acquiring accurate knowledge of the world (knowledge bottleneck), hand-coding this knowledge into a robot-accessible form (engineering bottleneck), computational intractability of optimally solving control problems in realistic settings (tractability bottleneck), and possible differences between the real world and the model of it used for planning (precision bottleneck). 
Many of these issues are still faced when developing and deploying AI systems for a variety of applications. 
\citet{ruvolo:13} focus on lifelong multi-task learning which alternates the training phase of each new task with testing phases for all previous tasks. They learn a library of latent model components that become a shared basis for all tasks which can be grouped or overlapped as needed.
Other lifelong learning systems aim to learn more general knowledge that is expected to be useful for a variety of tasks.
\citet{carlson:10} aim to use some initial annotated resources to a system that alternates between extracting facts for a knowledge base by machine reading on un-annotated documents, and using extracted facts to improve machine reading.
\citet{chen:iccv13} use Google Image Search to obtain data for training classifiers for objects, scenes and attributes, which is then used to learn visual relationships between objects, attributes and scenes. 
\citet{sun:icra16} propose an extension of the above which learns a hierarchy of object names using a combination of Bayesian modelling and crowdsourced annotations to more effectively learn classifiers for an open vocabulary of objects. Of these, only \citet{sun:icra16} explicitly address some of the difficulties around acquiring additional labels to continue learning. 
More focus on the intersection of dialog systems and lifelong learning, can help develop dialog acts and structures that can be used to collect such labelled data, address challenges such as potential user frustration, non-cooperation or misuse, and design methods to demonstrate this learning to users to encourage further cooperation. 

\section{Directions for Future Research}

While there is increasing interest in using dialog as a more general knowledge acquisition mechanism, further work needs to be done in a number of directions to make this a viable mechanism for practical applications.  
Existing work on opportunistic active learning, and more general reinforcement learning for active learning~\citep{thomason:corl17,padmakumar:emnlp18,fang:emnlp17} assume a cold start scenario, where all model learning must be done in the active learning phase. 
However, practical systems are likely to be pretrained on a reasonable amount of annotated data, with active learning primarily used for domain adaptation, or for handling novel concepts not seen at training time.  
Further work is required to demonstrate that these techniques are empirically useful when the system has to trade off model improvement with a mix of tasks that may or may not require model improvement.

Recent work has also highlighted some of the limitations of active learning in real-world settings -- with benefits not generalizing reliably across models and tasks, changing deployed models, as well as the model size, training data requirements and stochasticity of deep learning models~\cite{lowell:emnlp19,koshorek:conll19}. 
This suggests that new types of active learning methods may be required with deep learning models. 

Existing works that perform some form of lifelong learning through dialog typically do it either by asking the user to label examples for supervised learning, or by collecting facts to build a knowledge base that can then be used for downstream tasks. 
This has the potential to be extended to other types of information that can be useful for lifelong learning. 
For example, ~\citet{goyal:ijcai19} propose a method to convert natural language instructions to shaping rewards that can be used to speed-up learning policies in Atari games using Reinforcement Learning. 
A robot can potentially request instructions to achieve alternate goals in order to either improve its ability to understand such instructions, or to build a bank of policies expected to be useful in future interactions.
It could also request the human user to demonstrate alternate actions expected to be useful in the future. 
For example, if the user has asked the robot to perform a task that involves opening a can, the robot may ask for a demonstration of how to open the can if there was no can-opener available. 
Such a robot would need to be able to hallucinate an alternate goal that is expected to be useful, and determine whether it is appropriate to query for the demonstration of this goal both in the context of the dialog, and in the context of the original task the user requested. 

User frustration is another potential concern for frameworks such as opportunistic active learning. 
This restricts the number of queries that the system may ask per dialog session, which may require active learning methods to be combined with other semi-supervised learning techniques to scale up to the data requirements of deep learning models.
The spread of queries across dialogs in batches can also be improved using extensions of batch-mode active learning techniques~\citep{brinker:icml03,guo:nips08} to deep learning methods~\citep{ash:iclr20}.
In order to reduce user frustration due to active learning questions, another possibility would be to look into methods to implicitly embed active learning queries into system responses. For example, an interactive search and retrieval system that allows a user to refine search results can combine a mix of search results known to be relevant, with one or two results that it is uncertain about. Whether or not the user selects these can provide a weak, noisy label. 
In dialog systems where the user interacts via speech rather than text, additional cues can be used to decide when active learning queries are potentially inappropriate.
For example, prosodic cues can be used to identify whether users are stressed or frustrated~\citep{devillers:icphs03}, and the dialog policy can be designed to avoid active learning questions when such reactions are detected. Other sources of input such as face expressions~\citep{schuller:11} or gestures can also be used to improve such predictions~\cite{busso:08}.
Prosody may also be useful for detecting sarcasm~\citep{tepperman:icslp06,rakov:interspeech13} or other forms of misuse or intentional wrong answers from users, to avoid corrupting the collected labelled data. 

Depending on the types of active-learning questions involved, the systems may also need to demonstrate some sort of few-shot learning to keep users interested in assisting the learning process.
For example, children typically can learn to identify colors or common objects such as fruits with very few examples. 
For such tasks, users are likely to expect a similar rate of learning from the system. 
In contrast, more error may be tolerated in a system learning something less tangible such as a person's food preferences for the purpose of recommending restaurants.

Another challenge with learning dialog systems that include explicit queries for model improvement is the design of suitable simulation environments for learning effective dialog policies. 
A common trend for building dialog systems for a task or domain is to collect human-human dialogs as a starting point~\citep{budzianowski:emnlp18,de:cvpr17,thomason:corl19}.
However, if the concepts that the system is trying to learn are well know to human users, it is difficult for humans to ask questions that would be good active learning questions~\citep{yu:17}. 
Some work solves this problem by replacing the words denoting these concepts with words in a synthetic language~\citep{yu:17}.
However this idea is not easily adapted if the active learning queries need to obtain per-example labels for a concept, for example, asking whether an image contains a person or not. 
Other works make use of additional information available in the dataset, such as annotated attributes of objects~\citep{padmakumar:emnlp18} or a known navigation graph~\citep{chi:20,daume:emnlp19} that can potentially provide answers to any queries the system asks.
Simulation environments that require extensive extra annotation can be expensive to build, especially if these have to be task specific. 
If the dialog system has to make use of an existing annotated dataset, this restricts the set of information gathering actions to those that can be answered by the available annotations. 

This challenge becomes more significant for dialog systems that are intended to be a part of an embodied system. 
Most current work on dialog for embodied systems is done entirely in simulation~\citep{thomason:corl19,daume:emnlp19,padmakumar:emnlp18}, particularly if the work involves the learning of dialog policies. 
In addition to the difficulty of designing a simulation environment, further experiments are needed to evaluate whether existing systems perform comparably when implemented on real robots.
It is likely that additional work will be required in transferring dialog systems from simulated to real physical environments. 

\section{Conclusion}
Dialog systems have the potential to be important for lifelong learning, as a mechanism for collecting additional useful labelled data during operation. 
This is particularly relevant in the context of physically situated dialog systems such as robotic ones as data collection for these applications is more expensive and time-consuming.

The topic of adapting dialog systems for lifelong learning presents a number of interesting challenge problems for the dialog community including 
\begin{itemize}
    \item Designing dialog acts that can be used for collecting supervised data for various types of underlying systems -- an example would be designing a dialog act that allows a robot to simulate a goal it does not know how to reach and request a demonstration.
    \item Designing dialog acts that combine task oriented responses with queries for active learning -- for example, when asked to open a can, a robot can query whether it can be opened using a knife instead of a can opener.
    \item Prosodic analysis to identify urgency, stress, sarcasm and frustration in users to determine when it is appropriate to include or avoid active learning queries.
    \item Design of datasets or dialog simulation environments that can simulate changes in the environment over time, and provide answers to a wide range of information gathering actions.
    \item Dialog user studies to determine the effects of demonstrating evidence of learning.
\end{itemize}

\bibliography{acl2020}
\bibliographystyle{acl_natbib}

\end{document}